\documentclass[a4paper]{llncs}
\usepackage{amssymb}
\usepackage{graphicx}

\usepackage{url}
\urldef{\mailsa}\path|{alfred.hofmann, ursula.barth, ingrid.haas, frank.holzwarth,|
\urldef{\mailsb}\path|anna.kramer, leonie.kunz, christine.reiss, nicole.sator,|
\urldef{\mailsc}\path|erika.siebert-cole, peter.strasser, lncs}@springer.com|    
\newcommand{\keywords}[1]{\par\addvspace\baselineskip
\noindent\keywordname\enspace\ignorespaces#1}

\usepackage{times}
\usepackage{graphicx}
\usepackage{footnote}
\usepackage{array}
\usepackage{comment}
\usepackage{graphicx}
\usepackage{placeins}
\usepackage{float}
\usepackage{subfigure}
\usepackage[table,xcdraw]{xcolor}
\usepackage{url}
\usepackage{graphicx}
\usepackage{array}
\usepackage[inline]{enumitem}
\usepackage[table]{xcolor}
\usepackage{tabularx}
\usepackage{caption}
\usepackage{float}
\usepackage{tabu}
\usepackage{multicol}
\usepackage{multirow}
\usepackage{longtable}
\usepackage{flushend}
\usepackage{adjustbox}
\usepackage{anyfontsize}
\usepackage{footnote}
\makesavenoteenv{tabular}
\makesavenoteenv{table}
\usepackage{lipsum}
\usepackage{courier}
\usepackage{adjustbox}
\usepackage{booktabs} % http://ctan.org/pkg/booktabs

\usepackage{hyperref}
\colorlet{tableheadcolor}{gray!25} % Table header colour = 25% gray
\newcommand{\headcol}{\rowcolor{tableheadcolor}} %
\colorlet{tablerowcolor}{gray!10} % Table row separator colour = 10% gray
\newcommand{\rowcol}{\rowcolor{tablerowcolor}} %
\usepackage{booktabs}% http://ctan.org/pkg/booktabs
\usepackage{colortbl}% http://ctan.org/pkg/colortbl
\usepackage{amsmath}% http://ctan.org/pkg/amsmath
\usepackage{xcolor}% http://ctan.org/pkg/xcolor
\usepackage{graphicx}% http://ctan.org/pkg/graphicx
\colorlet{tableheadcolor}{gray!25} % Table header colour = 25% gray
\colorlet{tablerowcolor}{gray!10} % Table row separator colour = 10% gray
\newcommand{\topline}{\arrayrulecolor{black}\specialrule{0.1em}{\abovetopsep}{0pt}%
	\arrayrulecolor{tableheadcolor}\specialrule{\belowrulesep}{0pt}{0pt}%
	\arrayrulecolor{black}}
\newcommand{\midline}{\arrayrulecolor{tableheadcolor}\specialrule{\aboverulesep}{0pt}{0pt}%
	\arrayrulecolor{black}\specialrule{\lightrulewidth}{0pt}{0pt}%
	\arrayrulecolor{white}\specialrule{\belowrulesep}{0pt}{0pt}%
	\arrayrulecolor{black}}
\newcommand{\bottomline}{\arrayrulecolor{white}\specialrule{\aboverulesep}{0pt}{0pt}%
	\arrayrulecolor{black}\specialrule{\heavyrulewidth}{0pt}{\belowbottomsep}}%
\begin{document}

\mainmatter  

\title{Quality of Word Embeddings on Sentiment Analysis Tasks}

\titlerunning{Quality of Word Embeddings on Sentiment Analysis Tasks}

\author{Erion \c Cano%
\and Maurizio Morisio}

\authorrunning{Erion \c Cano and Maurisio Morisio}
\institute{Politecnico di Torino,\\
Duca degli Abruzzi, 24, 10129 Torino, Italy\\
\email{erion.cano@polito.it} \\
\email{maurizio.morisio@polito.it}
%\mailsa\\
%\mailsb\\
%\mailsc\\
%
}

\toctitle{Lecture Notes in Computer Science}
\tocauthor{Authors' Instructions}
\maketitle

\begin{abstract}
Word embeddings or distributed representations of words 
are being used in various applications like machine translation, 
sentiment analysis, topic identification etc. Quality of word embeddings 
and performance of their applications depends on several factors like training method, 
corpus size and relevance etc. In this study we compare performance of a dozen 
of pretrained word embedding models on lyrics sentiment analysis and movie review polarity
tasks. According to our results,  Twitter Tweets is the best on lyrics sentiment analysis, whereas 
Google News and Common Crawl are the top performers on movie polarity analysis. Glove trained models
slightly outrun those trained with Skip-gram. Also, factors like topic relevance and size of  
corpus significantly impact the quality of the models. When medium or large-sized 
text sets are available, obtaining word embeddings from same training dataset is usually the 
best choice.  
\end{abstract}
\section{Introduction}
Semantic vector space models of language were developed in the 90s to predict joint probabilities 
of words that appear together in a sequence.
A particular upturn was proposed by
Bengio et al. in \cite{Bengio:2003:NPL:944919.944966}, replacing sparse n-gram models with word embeddings 
which are more compact representations obtained using feed-forward or more advanced 
neural networks. Recently, high quality and easy to train Skip-gram shallow 
architectures were presented in \cite{DBLP:journals/corr/abs-1301-3781}and considerably 
improved in \cite{DBLP:conf/nips/MikolovSCCD13} with the introduction of negative 
sampling and subsampling of frequent words. 
The "magical" ability of word embeddings to capture syntactic and semantic regularities on text
words is applicable in various applications like machine translations, error correcting systems,
sentiment analyzers etc. This ability has been tested 
in \cite{linguistic-regularities-in-continuous-space-word-representations}
and other studies with analogy question tests of the form "A is to B as C is to \_\_" or male/female relations. 
A recent improved method for generating word embeddings is Glove \cite{conf/emnlp/PenningtonSM14}
which makes efficient use of global statistics of text words and preserves the linear substructure of 
Skip-gram word2vec, the other popular method.  
Authors report that Glove outperforms other methods such as Skip-gram in
several tasks like word similarity, word analogy etc. 
In this paper we examine the quality of word embeddings on 2 sentiment analysis tasks: 
Lyrics mood recognition and movie review polarity analysis. We compare various models pretrained with 
Glove and Skip-gram, together with corpora we train ourself. Our goal is to report the best performing 
models as well as to observe the impact that certain factors like training method, corpus size and thematic 
relevance of texts might have on model quality. 
According to the results, Common Crawl, Twitter Tweets and 
Google News are the best performing models. Corpus size and thematic relevance have 
a significant role on the performance of the generated word vectors. We noticed that models trained with 
Glove slightly outperform those trained with Skip-gram in most of experiments. 
\section{Word Embedding Corpora and Models}
In this section we present the different word embedding models that we compare. 
Most of them are pretrained and publicly available. Two of them (Text8Corpus and MoodyCorpus)
were trained by us. The full list with some basic characteristics is presented in Table 1. 
\begin{table}[ht] 
	\caption{List of word embedding corpora}  
    %\small 
%	\footnotesize
	\scriptsize
	%\tiny
	\centering      
	\setlength\tabcolsep{2.2pt}  
	\begin{tabular}
		{l c c c c c}  
		\topline
		\headcol \textbf{Corpus Name} & \textbf{Training} & \textbf{Dim} & \textbf{Size} & \textbf{Voc} & \textbf{URL}  	\\ [0.5ex] 
		\midline   
		Wiki Gigaword 300 & Glove & 300 & 6B & 400000 & \href{http://nlp.stanford.edu/projects/glove/}{link}			  \\
		\rowcol	Wiki Gigaword 200 & Glove & 200 & 6B & 400000 & \href{http://nlp.stanford.edu/projects/glove/}{link}			  \\
		Wiki Gigaword 100 & Glove & 100 & 6B & 400000 & \href{http://nlp.stanford.edu/projects/glove/}{link}			  \\
		\rowcol	Wiki Gigaword 50 & Glove & 50 & 6B & 400000 & \href{http://nlp.stanford.edu/projects/glove/}{link}			  \\
		Wiki Dependency & word2vec & 300 & 1B & 174000 &  %\footnotemark\ 
		\href{https://levyomer.wordpress.com/2014/04/25/dependency-based-word-embeddings/}{link} \\
		\rowcol Google News & word2vec & 300 & 100B & 3M & \href{https://code.google.com/archive/p/word2vec/}{link}			  \\ 
		Common Crawl 840 & Glove & 300 & 840B & 2.2M & \href{http://nlp.stanford.edu/projects/glove/}{link}			  \\
		\rowcol Common Crawl 42 & Glove & 300 & 42B & 1.9M & \href{http://nlp.stanford.edu/projects/glove/}{link}			  \\
		Twitter Tweets 200 & Glove & 200 & 27B & 1.2M & \href{http://nlp.stanford.edu/projects/glove/}{link}			  \\
		\rowcol Twitter Tweets 100 & Glove & 100 & 27B & 1.2M & \href{http://nlp.stanford.edu/projects/glove/}{link}			  \\
		Twitter Tweets 50 & Glove & 50 & 27B & 1.2M & \href{http://nlp.stanford.edu/projects/glove/}{link}			  \\
		\rowcol Twitter Tweets 25 & Glove & 25 & 27B & 1.2M & \href{http://nlp.stanford.edu/projects/glove/}{link}			  \\
		Text8Corpus & word2vec & 200 & 17M & 25000 & \href{https://cs.fit.edu/\%7Emmahoney/compression/textdata.html}{link}			  \\
		\rowcol MoodyCorpus & word2vec & 200 & 90M & 43000 & \href{http://softeng.polito.it/erion/}{link}			  \\
		\bottomline
	\end{tabular} 
\end{table}
\noindent Wikipedia Gigaword is a combination of Wikipedia 2014 dump and 
Gigaword 5
with 
about 6 billion tokens in total. It was created by authors of \cite{conf/emnlp/PenningtonSM14}
to evaluate Glove performance. 
Wikipedia Dependency corpus is a collection of 1 billion tokens from Wikipedia. The method 
used for training it is a modified version of Skip-gram word2vec  described in \cite{DBLP:conf/acl/LevyG14}. 
Google News is one of the biggest and richest text sets with 100 billion tokens and a vocabulary of 
3 million words and phrases \cite{DBLP:journals/corr/abs-1301-3781}. It 
was trained using Skip-gram word2vec with negative sampling, windows size 5 and 300 dimensions. 
Even bigger is Common Crawl 840,  a huge corpus of 840 billion tokens and 2.2
million word vectors also used at \cite{conf/emnlp/PenningtonSM14}. 
It contains data of Common Crawl (http://commoncrawl.org),
a nonprofit organization that 
creates and maintains public datasets by crawling the web. Common Crawl 42 is a reduced version 
made up of 42 billion tokens and a vocabulary of 1.9 million words. 
Common Crawl 840 and Common Crawl 42 were trained with Glove method producing vectors of 300
dimensions for each word. 
The last Glove corpus is the collection of Twitter Tweets. It consists of 2 billion tweets, 
27 billion tokens and 1.2 million words. 
To observe the role of corpus size in quality of  generated embeddings, we train and use Text8Corpus, a smaller corpus 
consisting of 17 million tokens and 25,000 words. 
The last model we use is MoodyCorpus, a collection of lyrics that followed our work 
in \cite{ismsi17} where we build and evaluate MoodyLyrics, a sentiment annotated dataset of songs. 
The biggest part of MoodyCorpus was built using lyrics of 
Million Song Dataset (MSD) songs (https://labrosa.ee.columbia.edu/millionsong/).
As music tastes and characteristics change over 
time (http://kaylinwalker.com/50-years-of-pop-music), 
it is better to have diversified sources of songs in terms of epoch, genre etc. Thereby 
we added songs of different genres and 
epochs that we found in  
two subsets of MSD, 
Cal500 
and TheBeatles. 
The resulting corpus of 90 million tokens and 43,000 words can be 
downloaded from http://softeng.polito.it/erion. Further information about public music 
datasets can be found at \cite{7325106}.  
\section{Sentiment Analysis Tasks}
The problem of music mood recognition is about utilizing 
machine learning, data mining and other techniques to automatically classify songs in 
2 or more emotion categories with highest possible accuracy. Different combinations of features 
such as audio or lyrics are involved in the process. 
In this study we make use of song lyrics exploiting the dataset described in \cite{7536113} (here AM628). 
The original dataset contains 771 song texts collected from AllMusic
portal. AllMusic tags and 3 human experts were used for the annotation of songs. We balanced the dataset 
obtaining 314 positive and 314 negative lyrics. 
We also utilize MoodyLyrics (here ML3K),  a dataset of 3,000 mood labeled songs from different genres 
and epochs described in \cite{ismsi17}. 
Pioneering work in movie review polarity analysis has been conducted by Pang 
and Lee in \cite{Pang+Lee+Vaithyanathan:02a} and \cite{Pang+Lee:04a}. 
The authors released sentiment polarity dataset, a collection of 2,000 movie reviews 
categorized as positive or negative. 
Deep learning techniques and distributed word representations appeared on recent studies like
\cite{shirani2014applications} where the role of RNNs (Recurrent Neural Networks), 
and CNNs (Convolutional Neural Networks) is explored. The author reports 
that CNNs perform best.
An important work that has relevance here is \cite{maas-EtAl:2011:ACL-HLT2011} where authors 
present 
an even larger movie review dataset of 50,000 
movie reviews from IMBD. 
This dataset has been used in various works 
such as \cite{DBLP:conf/naacl/Johnson015}, \cite{pouransari2014deep} etc. For our experiments we 
used a chunk of 10K (MR10K) as well as the full set (MR50K).
We first cleaned and tokenized texts of the datasets. 
The dataset of the current run is loaded and a set of unique text words is created. 
All 14 models are also loaded in the script. We train a 15th 
(self\_w2v) model using the corpus of the current run and Skip-gram method. 
The script iterates in every line of the pretrained models splitting apart the words and the 
float vectors and building \{word: vec\} dictionaries later used as classification feature sets.
Next we prepare the classification models using tf-idf vectorizer which has been successfully 
applied in similar studies like \cite{DBLP:conf/ismir/HuDE09}.
Instead of applying tf-idf in words only as in other text classifiers, we vectorize both 
word (for semantic relevance) and corresponding vector (for syntactic and contextual relevance). 
Random forest was used as classifier and 5-fold cross-validation accuracy is computed for each of 
the models. 
\section{Results}
In figures 1 and 2 we see results of 5-fold cross-validation on the 2 lyrics datasets.
Top three models are crawl\_840, twitter\_50 
and self\_w2v. On AM628 (very smallest dataset), it is crawl\_840 (the biggest model) that leads, 
followed by twitter\_50.  Self\_w2v is severely penalized by its size and thus is at the bottom.  
On ML3K (large dataset) self\_w2v reaches the top of the list, leaving behind twitter\_50. 
Wikigiga, google\_news and dep\_based are positioned in the middle whereas 
MoodyCorpus and Text8Corpus end the list. Their accuracy scores drift from 0.62 to 0.75. 
It is interesting to see how self\_w2v goes up from the last to the top, with scores edging between 
0.61 and 0.83. This model is trained with the data of each experiment and depends on the size 
of that dataset which grows significantly (see Table 2). 
We see that accuracy values we got here are in line with reports from other similar works 
such as 
\cite{10.1109/ISM.2009.123} where they use a dataset of 1032 lyrics from 
AllMusic to perform content analysis with text features. 
\begin{figure}[!t]
	\centering
	\begin{minipage}{.51\columnwidth}
		\centering
		\includegraphics[width=0.97\columnwidth]{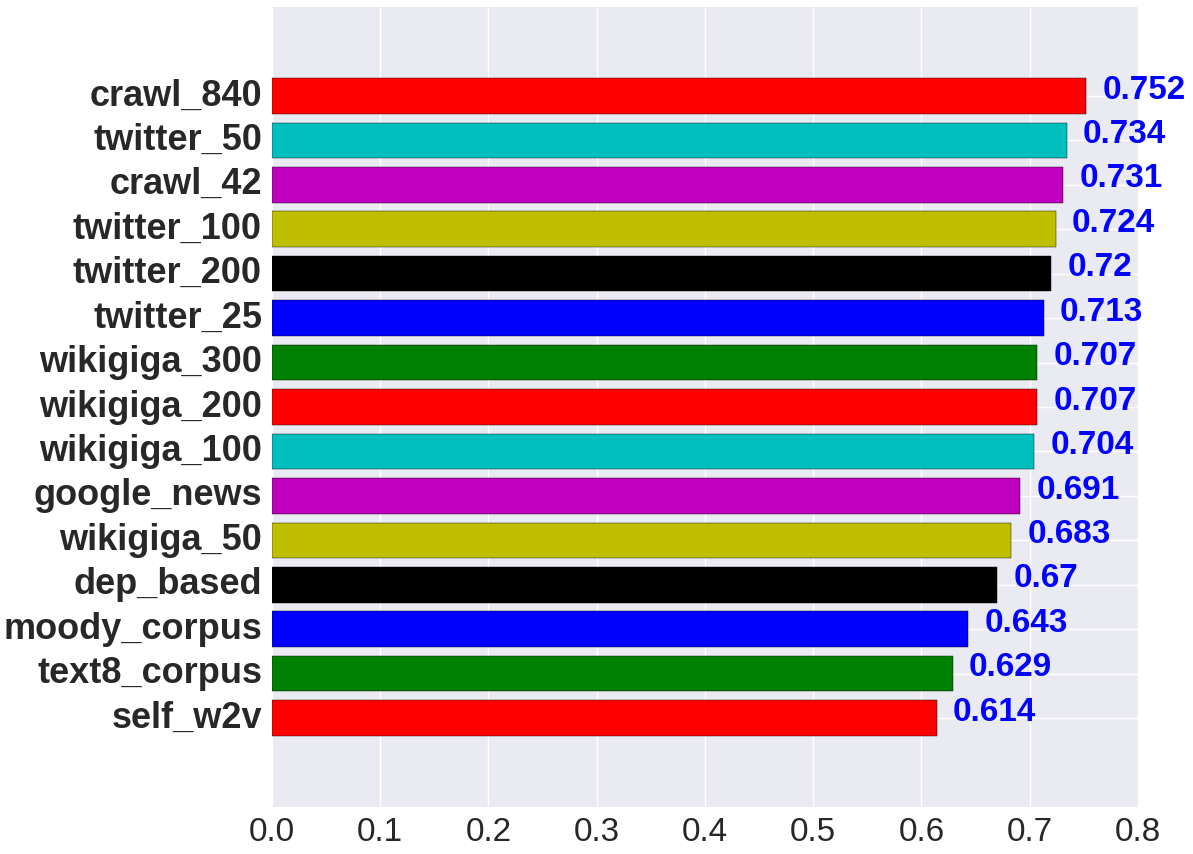}
		\caption{Lyric accuracies on AM628 }
		\label{fig:mockup1}
	\end{minipage}%
	\begin{minipage}{.51\columnwidth}
		\centering
		\includegraphics[width=0.97\columnwidth]{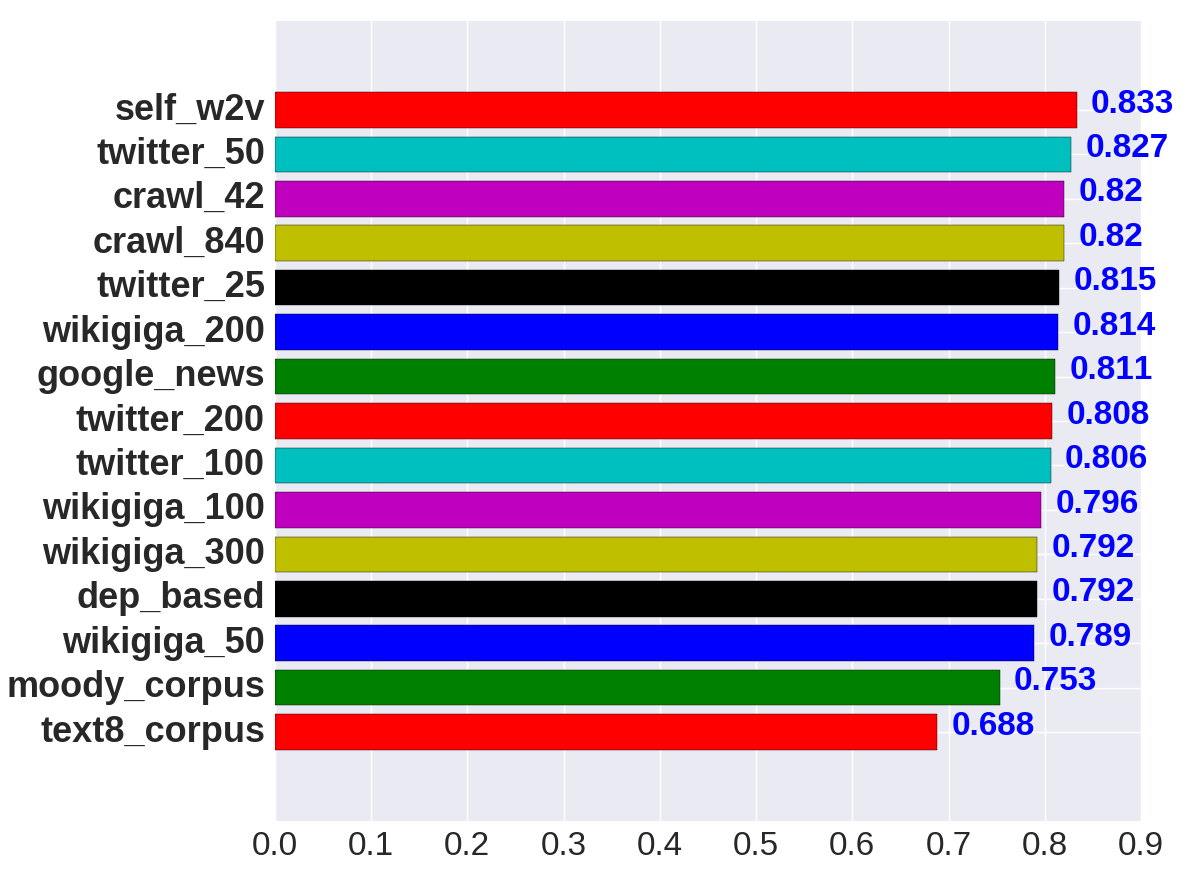}
		\caption{Lyric accuracies on ML3K}
		\label{fig:mockup2}
	\end{minipage}
\end{figure}
Accuracy scores for movie review polarity prediction are presented in figures 3 and 4.
\begin{figure}[!t]
	\centering
	\begin{minipage}{.51\columnwidth}
		\centering
		\includegraphics[width=0.97\columnwidth]{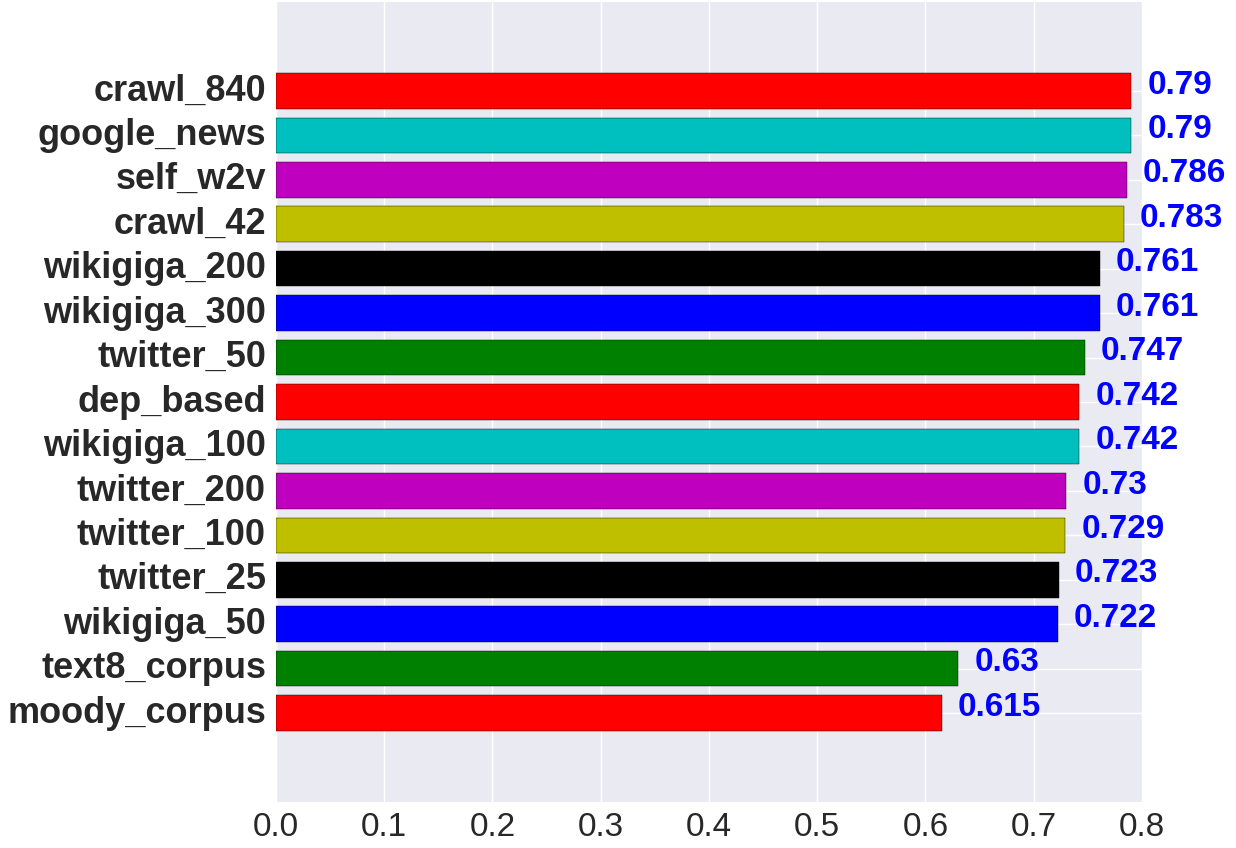}
		\caption{Review accuracies on MR10K}
		\label{fig:mockup2}
	\end{minipage}%
	\begin{minipage}{.51\columnwidth}
		\centering
		\includegraphics[width=0.97\columnwidth]{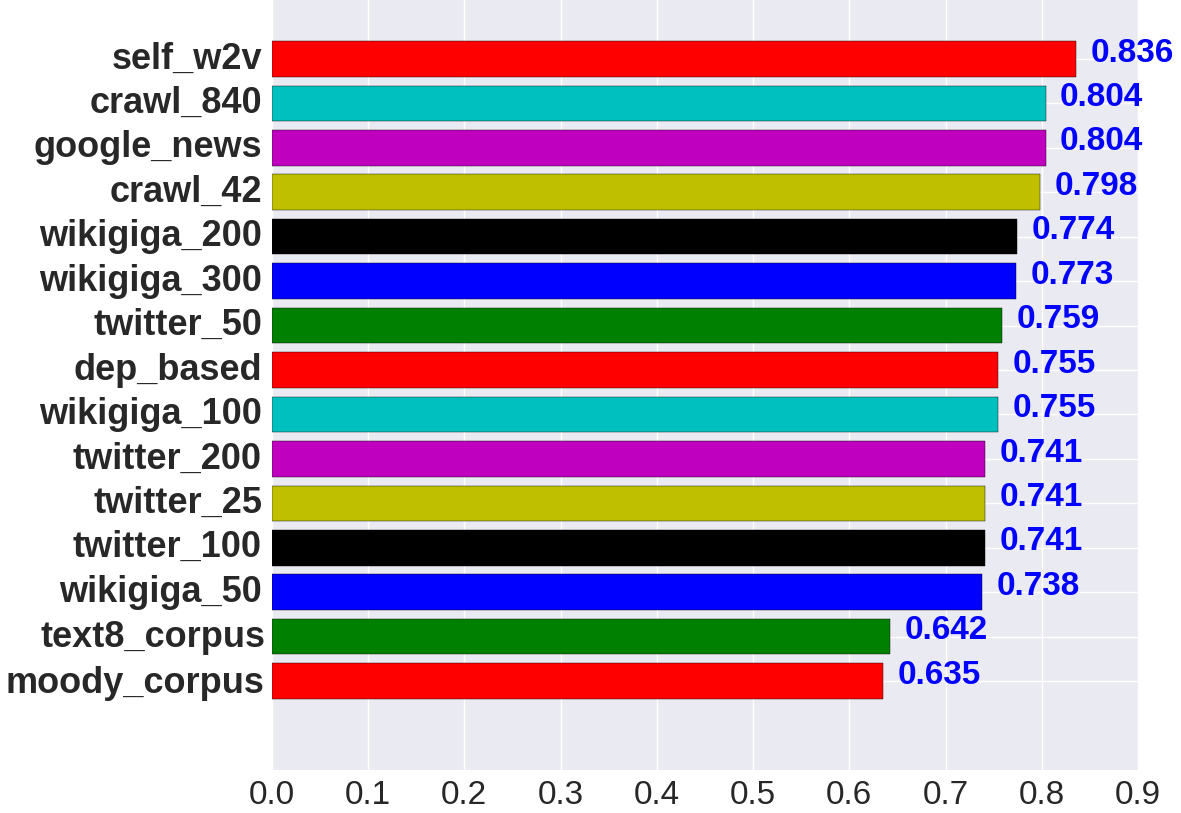}
		\caption{Review accuracies on MR50K}
		\label{fig:mockup2}
	\end{minipage}
\end{figure}
Again we see that crawl\_840 performs very well. Google\_news is also among the top 
whereas Twitter models are positioned in the middle of the list. Once again self\_w2v grows considerably, 
this time from the 3rd place to the top. On MR50K it has a discrete margin of more than 0.03
from the 2nd position. Again wikigiga models are positioned in the middle of the list and the worst 
performing models are MoodyCorpus and Text8Corpus. 
Our scores on this task are somehow lower than those reported from various studies that 
explore advanced deep learning constructs  on same dataset. 
In \cite{maas-EtAl:2011:ACL-HLT2011} for example, authors who created movie review dataset try on it their 
probabilistic model that is able to capture semantic similarities between words. They report a maximal accuracy 
of 0.88. 
A study that uses 
a very similar method is \cite{pouransari2014deep} where authors combine random forest with word vector average
values. On movie review dataset they achieve an accuracy of 0.84 which is about what we got here.
\begin{table}[ht] 
	\caption{Properties of self\_w2v}  
%	\small 
%	\footnotesize
	\scriptsize
%  \tiny
	\centering    
	\setlength\tabcolsep{2pt}  
	\begin{tabular}
		{c c c c c c}  
		\topline
		\headcol \textbf{Trial} & \textbf{Dataset} & \textbf{Dim} & \textbf{Size} & \textbf{Voc} & \textbf{Score}		 	\\ [0.5ex] 
		\midline   
		1 & AM628 & 200 & 156699 & 8756	& 0.614	  \\
%		\rowcol	2 & ML1K & 200 & 344499& 11012 & 0.749		  \\
%		3 & ML2K & 200 & 685593 & 14142	& 0.806	  \\
		\rowcol	2 & ML3K & 200 & 1028891 & 17890 & 0.833		  \\
%		\hline
%		5 & MR5K & 300 & 1179453 & 39518 & 0.739	 \\
		3 & MR10K & 300 & 2343641 & 53437 & 0.786	  \\ 
%		7 & MR15K & 300 & 3523274	 & 63333 & 0.799		  \\
		\rowcol 4 & MR50K & 300 & 11772959 & 104203	& 0.836	  \\
		\bottomline
	\end{tabular} 
\end{table}
\section{Discussion}
In this paper we examined the quality of different word embedding models on two sentiment 
analysis tasks: Lyrics mood recognition and movie review polarity. We observed
the role of factors like training method, vocabulary and corpus size and thematic 
relevance of texts. 
According to our results, the best performing models are Common Crawl, Twitter Tweets and 
Google News. In general,
models trained with Glove slightly outperform those trained using Skip-gram, especially on 
lyrics sentiment analysis (Twitter and Crawl). 
We also notice that vocabulary richness and corpus size have a significant influence on model quality. The biggest 
models like crawl\_840 are always among the best. Likewise self\_w2v performs very well on both tasks
when trained with medium or large data sizes (see Table 2). 
Being the smallest in sizes, MoodyCorpus and Text8Corpus are always the worst. 
Regarding thematic relevance, 
Twitter corpora perform better on lyrics sentiment analysis. 
They are large and rich in 
vocabulary with texts of an informal and sentimental language. This language is very similar 
to the one of song lyrics, with \emph{\textbf{love}} being the predominant 
word (see word cloud in \cite{ismsi17}). 
Movie review results on the other hand, are headed by Common Crawl and Google News which are 
the largest, both in size and vocabulary. These models are trained with diverse and informative texts 
that cover every possible subject or topic. Having a look on some movie 
reviews we also see a similar language with comments about the movies of different categories. 
Furthermore, 
we saw that when training set is big enough, obtaining word embeddings from it (self\_w2v) is 
the best option.
%
%
%% bibtex bibliography
\bibliographystyle{abbrv}
\bibliography{wordemb}  
\end{document}